\documentclass[onecolumn,british,twocolumn ]{IEEEtran}
\usepackage[T1]{fontenc}
\usepackage[latin9]{inputenc}
\usepackage{amsmath}
\usepackage{amssymb}
\usepackage{graphicx}

\makeatletter

\newcommand*\LyXZeroWidthSpace{\hspace{0pt}}

\usepackage{hyperref}

\makeatother

\usepackage{babel}
\begin{document}
\title{The Silent Scholar Problem: A Probabilistic Framework for Breaking
Epistemic Asymmetry in LLM Agents}
\author{%
\begin{minipage}[t]{0.3\textwidth}%
\begin{center}
Zan-Kai Chong\\
\emph{School of Science and\\ Technology\\
Kwansei Gakuin University}\\
Japan\\
zankai@ieee.org
\par\end{center}%
\end{minipage}%
\begin{minipage}[t]{0.3\textwidth}%
\begin{center}
Hiroyuki Ohsaki\\
\emph{School of Science and\\ Technology\\
Kwansei Gakuin University}\\
Japan\\
ohsaki@kwansei.ac.jp
\par\end{center}%
\end{minipage}%
\begin{minipage}[t]{0.3\textwidth}%
\begin{center}
Bryan Ng\\
\emph{ 
School of Engineering \& Computer Science\\
Victoria University of Wellington}\\
New Zealand
\\
ckbryan@hotmail.com
\par\end{center}%
\end{minipage}}
\maketitle
\begin{abstract}
Autonomous agents powered by large language models (LLMs), often enhanced
by retrieval-augmented generation (RAG), are powerful processors of
digital content. However, RAG is limited to consumption, constraining
agents as unidirectional consumers. We term this limitation epistemic
asymmetry. This imbalance creates a globally inefficient landscape,
where silent scholars redundantly reconstruct similar reasoning in
isolation, stagnating collective intelligence. To break this asymmetry,
we propose a formal probabilistic framework that provides agents with
a non-altruistic motive to engage in bidirectional knowledge exchange.
We model an agent's belief in a proposition as a Beta-Bernoulli distribution
with a forgetting factor ($\gamma$). This allows us to isolate epistemic
uncertainty as the variance of belief, establishing a dual motive
for interaction: (1) a homeostatic drive to maintain certainty against
temporal decay, and (2) an optimal learning strategy that targets
maximum ambiguity ($\text{E}[\theta]=0.5$). Furthermore, we address
the practical challenges of deployment by introducing epistemic caching,
a mechanism that leverages the forgetting factor to dynamically allocate
resources to the non-stationary active head of the knowledge distribution.
While our current analysis focuses on independent propositions and
binary feedback to establish a baseline for this motive, we discuss
how the framework extends to correlated and multi-categorical knowledge
structures. Finally, we discuss how these accumulated belief states
can serve as verifiable reward signals for reinforcement learning
from human feedback (RLHF) and high-quality data filters for supervised
fine-tuning (SFT), effectively bridging the gap between inference-time
interaction and long-term model alignment. Simulation results validate
this architecture, demonstrating that our uncertainty-driven strategy
significantly outperforms random baselines in heterogeneous (Zipfian)
environments, successfully prioritising the active head of the knowledge
distribution while maintaining adaptability to concept drift.
\end{abstract}

\section{Introduction}

The emergence of large language models (LLMs) has ushered in a new
generation of autonomous agent systems \cite{wang2024survey}. These
agents exhibit exceptional proficiency in analysing extensive information
from the internet, performing intricate multi-step operations, and
integrating knowledge to respond to user queries \cite{yao2022react}.
Ranging from advanced personal assistants to automated research platforms,
they mark a major advancement in artificial intelligence, serving
as highly capable processors and interpreters of digital content.

However, despite their breadth of capabilities, these agents remain
constrained by the static nature of their pre-trained knowledge. Retrieval-augmented
generation (RAG) has emerged as a significant step toward addressing
this, enabling LLMs to query external web resources to enhance factual
accuracy and mitigate hallucination \cite{lewis2021retrievalaugmentedgenerationknowledgeintensivenlp,gao2024retrievalaugmentedgenerationlargelanguage}.
Often, the retrieved documents provide valuable, context-specific
examples that the agent can use for in-context few-shot learning to
improve its immediate response.

Yet, RAG, even when used for few-shot learning, only solves the problem
of consumption. Contemporary agents remain architecturally constrained
as unidirectional consumers of knowledge, with minimal mechanisms
for reintegrating their synthesised insights into shared digital ecosystems.
We identify this limitation as epistemic asymmetry, and it creates
a critical failure mode for the agent itself.

Without bidirectional exchange, an agent cannot distinguish between
aleatoric noise (environmental randomness) and epistemic ignorance
(model deficiency), terms coined in \cite{wang2025aleatoricepistemicexploringuncertainty}.
Isolated from external correction, the agent is forced to train on
its own unverified outputs, a recursive process that leads to model
collapse. In this degenerative state, the agent\textquoteright s belief
distribution loses variance, and it becomes confidently wrong and
incapable of adapting to concept drift \cite{Shumailov2024ModelCollapse}.

Current agents lack the motive to break their asymmetry because they
lack a formal model to quantify this reducible, epistemic uncertainty.
Without a formal uncertainty model, an agent cannot distinguish between
environmental noise (which it should ignore) and its own ignorance
(which it should resolve). Consequently, they have no mechanism to
understand why contributing back to the digital commons (e.g., public
forums, Q\&A sites, Stack Overflow, Reddit, etc.) would be beneficial.
For an agent, posting a solution to a public problem and receiving
feedback is a powerful, low-cost method to gain new data. This feedback
is the very evidence needed to reduce its epistemic uncertainty, but
current models are not equipped to quantify this value.

In this paper, we envision that the contemporary agents will evolve
from silent scholars to becoming epistemic agents and actively engaging
in bidirectional knowledge exchange. To ground this vision, we propose
the formal probabilistic framework that provides their non-altruistic
motive. We initially model the agent\textquoteright s belief over
its propositions as an unknown success rate using a Beta-Bernoulli
model with a forgetting factor ($\gamma$). While this treats propositions
as independent units of evidence, it provides a necessary simplification
for our initial derivation of the equilibrium sample size ($N_{\text{eq}}$)
and the mathematical foundation for more complex, interdependent knowledge
representations.

This framework provides a non-altruistic motive: First, the forgetting
factor ensures that certainty decays over time, effectively converting
stale knowledge back into epistemic uncertainty. This prevents the
variance from decaying to zero, establishing a persistent motive for
continuous engagement. Second, our analysis indicates that the agent's
potential for learning is mathematically maximised at the point of
highest ambiguity ($E[\theta]=0.5$). These findings provide the formally-justified
motive for an agent to break its asymmetry. Furthermore, we address
the challenge of scalability by deriving an eviction policy that functions
as epistemic caching. By leveraging the forgetting factor, the agent
creates a dynamic working set of beliefs, ensuring the model remains
computationally tractable despite the vast proposition space of real-world
LLMs.

The remainder of this paper is organised as follows. Section \ref{sec:Related-Work}
reviews related work in autonomous agents and active learning , and
explicitly positions our framework against established paradigms such
as online Bayesian updating and multi-armed bandits. Section \ref{sec:Mathematical-Model}
formalises our probabilistic framework, detailing the Beta-Bernoulli
model used to represent an agent's belief state and uncertainty. Then
it provides a formal analysis of this model's properties, establishing
a non-altruistic motive of persistent uncertainty and maximum ambiguity.
Section \ref{sec:Simulation-Results} presents our experimental setup
and results validating this framework. Section \ref{sec:Discussion}
discusses the broader implications for system robustness and model
alignment, proposing mechanisms to mitigate re-calibration latency
and distill external belief states into the model's intrinsic weights
via SFT and RLHF. Finally, Section \ref{sec:Conclusion} concludes
our work.

\section{Related Work\label{sec:Related-Work}}

Our work is situated at the nexus of several rapidly evolving fields.
To motivate our contribution, we first review the capabilities and
limitations of existing agent architectures, focusing on their role
as unidirectional knowledge consumers. We then analyse the isolated
learner paradigm of current self-reflection frameworks. Finally, we
situate our probabilistic model as a novel bridge between the heuristic
learning of agents and the formal principles of active learning.

\subsection{Autonomous Agents}

The advent of LLMs has catalysed a paradigm shift in autonomous systems.
As extensively documented in recent surveys, these models form the
cognitive core of a new class of agents capable of complex reasoning,
planning, and tool use \cite{wang2024survey}. Agents powered by LLMs
can decompose high-level instructions into executable steps, interact
with external APIs, and synthesise information to achieve sophisticated
goals. This body of work has established the foundation for agents
that can effectively function as information processors and task executors. 

Recently, agent development has increasingly focused on deeper integration
with both web-based and desktop environments, enabling more seamless
user experiences and greater task automation. Instead of relying solely
on APIs, a new generation of browser agents interacts directly with
graphical user interfaces to perform tasks. Tech companies like Perplexity,
Anthropic, and OpenAI have developed agents that can navigate websites,
fill out forms, manage calendars, and synthesise information across
multiple open tabs \cite{openai_cua_2025,anthropic_claude_chrome_2025,perplexity_comet_2025}.
These systems demonstrate a clear trajectory towards agents that are
not just connected to the internet, but are situated within it, using
the web as their native workspace. This evolution validates the internet
as a viable environment for complex agentic behaviour. Nonetheless,
these agents are still architecturally designed as advanced assistants
that act on behalf of a single user, consuming and processing information
without a mechanism for reciprocal knowledge contribution.

\subsection{Agent Learning and Reflection}

A more advanced class of agents attempts to learn from their own operational
experience through internal feedback loops. A prominent example is
the Reflexion framework \cite{shinn2023reflexionlanguageagentsverbal},
which enables an agent to perform self-reflection on its past failures.
By analysing its own action history, the agent can generate verbal
feedback for itself, which it then uses to improve its performance
on subsequent trials. This concept of learning from a history of feedback
is also formalised in frameworks like Chain of Hindsight (CoH) \cite{liu2023chainhindsightalignslanguage},
which fine-tunes a model on its past trajectories.

This trend of internal self-evaluation extends beyond just past failures.
Tree of Thoughts (ToT) \cite{yao2023treethoughtsdeliberateproblem},
for example, allows an agent to deliberately explore and heuristically
evaluate multiple reasoning paths concurrently. Similarly, Self-RAG
\cite{asai2023selfraglearningretrievegenerate} introduces reflection
tokens that allow an agent to critique its own retrieval and generation
steps in real-time. Collectively, these frameworks represent a significant
step beyond static agents by endowing them with the capacity for self-correction.

This entire line of research, however, reveals two critical gaps.
First, the learning process, whether it is reflection (Reflexion,
CoH), deliberation (ToT), or self-critique (Self-RAG), remains a fundamentally
private and isolated activity. An agent's insights are not shared,
and the experience it gains is not contributed back to any external
ecosystem, thereby reinforcing its epistemic isolation.

Second, the learning process itself is largely heuristic. The decision
to reflect, backtrack, or critique is often triggered by an LLM's
self-generated score, simple failure detection, or a non-probabilistic
heuristic. This literature lacks a formal, mathematical definition
of an agent's belief state or its certainty. This makes it difficult
to quantify why an agent should seek external feedback or when it
has learned enough, a gap our probabilistic model directly addresses.

\subsection{Active Learning and Uncertainty Sampling}

To contextualize active learning, it is necessary to distinguish between
the types of uncertainty an agent encounters. As reviewed by Wang
et al. \cite{wang2025aleatoricepistemicexploringuncertainty}, uncertainty
is categorised into aleatoric (inherent data noise, irreducible) and
epistemic (model ignorance, reducible). Active learning strategies
are fundamentally designed to target epistemic uncertainty, as this
is the only component that can be reduced through the acquisition
of new data.

The field of active learning provides the formal mathematical tools
to address the shortcomings of heuristic learning \cite{liu2023understandinguncertaintysampling}.
Active learning is a subfield of machine learning where a learning
algorithm can intelligently choose the data from which it learns.
The goal is to achieve higher accuracy with fewer labelled examples,
and thus maximising learning efficiency.

The principles of active learning are well-established, but their
application has been almost exclusively limited to a model-dataset
paradigm. In this traditional view, an agent actively queries a static,
pooled dataset to request labels for the most informative unlabelled
points. This literature has not, to our knowledge, framed uncertainty
as a social or economic motive for an agent to engage in a public,
bidirectional exchange. Furthermore, a fundamental divergence exists
in the objective of sampling. Traditional active learning seeks to
minimise the query budget required to reach a static convergence point
(i.e., sampling stops when uncertainty $\approx0$). In contrast,
our approach views uncertainty not as a metric to be eliminated, but
as a persistent driver of existence. By introducing a forgetting factor,
we ensure the agent never converges to absolute certainty, thereby
transforming active learning from a finite optimisation task into
a continuous, lifelong homeostatic process.

Our work bridges this critical gap. We propose that the formal principles
of uncertainty sampling can be repurposed as the non-altruistic, formal
justification for an epistemic agent to break its asymmetry. We reframe
\textquotedbl querying a static dataset\textquotedbl{} as \textquotedbl posting
a solution on a public forum\textquotedbl{} and \textquotedbl receiving
a label\textquotedbl{} as \textquotedbl gaining feedback from the
digital commons.\textquotedbl{} In doing so, we provide the formal,
probabilistic motive that is missing from current agent architectures.

\subsection{Theoretical Positioning}

While our framework leverages established mathematical components,
specifically Beta-Bernoulli updating, exponential forgetting, and
uncertainty sampling, our contribution lies in the architectural synthesis
of these tools to resolve the silent scholar problem. We explicitly
distinguish our approach from prior work in four key areas:
\begin{enumerate}
\item Online Bayesian Updating \& Concept Drift: Traditional online learning
aims to passively track evolving parameters or detect drift to trigger
retraining \cite{gama2014survey}. In contrast, we deploy the forgetting
factor ($\gamma$) not as a tracking mechanism, but as a homeostatic
driver. The decay does not merely reflect a changing world, it actively
compels the agent to engage with the digital commons to prevent its
internal belief state from degenerating into maximum entropy.
\item Multi-Armed Bandits: Bandit algorithms (e.g., Thompson Sampling) typically
optimise for cumulative external reward, balancing exploration and
exploitation \cite{russo2020tutorialthompsonsampling}. Our framework
is distinct in that it is non-altruistic and reward-agnostic. In other
words, the agent optimises solely for epistemic maintenance. It treats
variance reduction not as a means to a reward, but as the survival
objective itself.
\item RAG Caching: Standard RAG caching stores static retrieved documents
or vector embeddings to reduce latency or cost \cite{Bergman_2025}.
Our epistemic caching (Section \ref{subsec:Scalability-via-Epistemic})
stores dynamic belief states $(\alpha,\beta)$. This allows the agent
to maintain a living model of truth that evolves with every interaction,
rather than a static snapshot of text.
\end{enumerate}

\section{Mathematical Model \label{sec:Mathematical-Model}}

In this section, we model the non-altruistic motive of epistemic agents
in verifying a portfolio of propositions. We employ a Beta-Bernoulli
model with a forgetting factor ($\gamma$) to capture the evolving
belief associated with each proposition. Building on this, we introduce
a formal definition of uncertainty that supports query strategy and
analyse the key properties of these dynamic models, including the
adaptability-certainty trade-off. Finally, we demonstrate how this
framework scales to real-world environments by introducing epistemic
caching, a mechanism that leverages the forgetting factor to prioritize
the retention of structurally significant knowledge while evicting
obsolete beliefs.

\subsection{Dynamic Belief Representation with Beta-Bernoulli Model with Forgetting
Factor\label{subsec:beta-gamma-tracker}}

We model the agent's knowledge base as a portfolio of $k$ propositions,
$P=\{p_{1},p_{2},...,p_{k}\}$. These propositions represent facts,
claims, or the posited effectiveness of a reasoning method. While
the theoretical space of propositions is vast, we treat $P$ as a
dynamic working set. As we will detail in Section \ref{subsec:Scalability-via-Epistemic},
the model naturally supports dynamic allocation to handle the long-tail
distribution of propositions found in real-world LLMs, allowing the
agent to focus only on the active head of the distribution. Consequently,
for the formal analysis in Sections \ref{subsec:beta-gamma-tracker}
through \ref{subsec:dynamic-model-properties}, we treat $P$ as the
finite set of currently active propositions.

For each proposition $p_{i}$, the agent's goal is to learn and track
its current probability of being true, $\theta_{i}(t)\in[0,1]$ at
discrete time step $t$ using a Beta-Bernoulli model. Correspondingly,
the agent's belief for proposition $p$ is represented by a $\text{Beta}(\alpha_{i},\beta_{i})$
distribution, where $\alpha_{i}$ and $\beta_{i}$ are pseudo-counts
for positive and negative evidence, respectively.

The agent's environment is the dynamic, non-stationary digital commons
(the internet), where the public consensus or supporting evidence
for a proposition can evolve. Critically, to operate in a non-stationary
environment, there is no basis for assuming that a proposition\textquoteright s
truth value remains constant. Public consensus may shift, or new evidence
may invalidate old facts. Therefore, the agent must prioritise recent
data over stale, historical observations. To address this, we introduce
a forgetting factor, $\gamma$, where $0\ll\gamma<1$ (e.g., $\gamma=0.99$),
to achieve this.

This factor exponentially decays the weight of old observations while
incorporating new evidence. We denote the observation at time $t$
as a binary variable $y_{t}\in\{0,1\}$, where $y_{t}=1$ represents
evidence supporting the proposition and $y_{t}=0$ represents contradictory
evidence. The belief update rule is defined as,
\begin{equation}
\alpha_{t}=\gamma\alpha_{t-1}+y_{t},\label{eq:alpha-update}
\end{equation}
and
\begin{equation}
\beta_{t}=\gamma\beta_{t-1}+(1-y_{t}).\label{eq:beta-update}
\end{equation}

While we define $y_{t}\in\{0,1\}$ as a binary evidence signal for
the purposes of this foundational model, the Beta-Bernoulli update
rule is mathematically compatible with soft evidence. In scenarios
where feedback is nuanced, $y_{t}$ can be treated as a continuous
variable $y_{t}\in[0,1]$, where values closer to $1$ represent stronger
supporting evidence and values closer to $0$ represent contradictory
evidence. This allows the framework to process probabilistic feedback
from critic models or human-in-the-loop signals without loss of generality.

\subsection{Uncertainty as the Motive for Interaction\label{subsec:uncertainty-motive}}

We quantify the agent's epistemic uncertainty as the variance of the
belief distribution for a specific proposition $p_{i}$. While this
study treats propositions as independent units to derive the equilibrium
sample size $N_{\text{eq}}$, this individual variance serves as the
base component for more complex, joint-uncertainty models in correlated
knowledge structures

The agent's primary challenge is one of active learning: distinguishing
between environmental noise and knowledge gaps. We specifically model
the agent's epistemic uncertainty regarding proposition $p_{i}$.
Unlike aleatoric uncertainty, which arises from the intrinsic randomness
of the environment and is irreducible, epistemic uncertainty stems
from the model's lack of knowledge regarding the true parameter $\theta$.

We quantify this epistemic uncertainty as the variance of the belief
distribution, i.e.,

\begin{equation}
\text{\text{Var}(\ensuremath{\theta})}=\frac{\alpha\beta}{\left(\alpha+\beta\right)^{2}\left(\alpha+\beta+1\right)}.\label{eq:uncertainty-1}
\end{equation}

By maximising this variance, the agent targets the region of highest
ambiguity, where the potential for information gain and thus the reduction
of epistemic uncertainty is maximised.

\subsection{Key Properties of the Dynamic Model\label{subsec:dynamic-model-properties}}

This Beta-Bernoulli model with forgetting factor provides the following
three properties. The agent's non-altruistic drive to reduce its own
uncertainty compels it to continuously interact with the digital commons
(see Section \ref{subsec:persistent-uncertainty}) and to intelligently
prioritise those interactions on the topics of its greatest confusion
(see Section\ref{subsec:maximum-ambiguity}). Then, the $\gamma$
serves as the key to adjust the adaptability in Section \ref{subsec:Adaptability-Certainty-Trade}.

\subsubsection{Persistent Uncertainty\label{subsec:persistent-uncertainty}}

In a static model (where $\gamma=1$), with $N=\alpha+\beta$ as $N\rightarrow\infty$,
the denominator of Eq. \ref{eq:uncertainty-1} will grow faster than
the numerator. Hence, causing $\text{Var}(\theta)\rightarrow0$. 

However, this is no longer true in the dynamic model. We define the
effective sample size as the sum of belief counts, $N_{\text{eff},t}=\alpha_{t}+\beta_{t}$.
By summing the update rules in Eq. \ref{eq:alpha-update} and \ref{eq:beta-update},
we derive the recurrence relation for the agent's memory,
\begin{align}
N_{\text{eff},t} & =(\gamma\alpha_{t-1}+y_{t})+(\gamma\beta_{t-1}+(1-y_{t}))\nonumber \\
 & =\gamma\left(\alpha_{t-1}+\beta_{t-1}\right)+1\nonumber \\
N_{\text{eff},t} & =\gamma N_{\text{eff},t-1}+1.
\end{align}

As the time progresses, an equilibrium point ($N_{\text{eq}}$) is
found, where $N_{\text{eff}}$ stops changing. This happens when the
amount of memory we lose to decay is exactly equal to the +1 we gain
from new evidence. At steady state, $N_{\text{eq}}=\left(N_{\text{eq}}\times\gamma\right)+1$
and 
\begin{equation}
N_{\text{eq}}=\frac{1}{1-\gamma}.
\end{equation}

Conclusively, the agent's effective sample size ($N_{\text{eff}}$)
stabilises instead of growing to infinity. As $N_{\text{eff}}$ is
capped, the denominator in Eq. \ref{eq:uncertainty-1} stops growing.
This leads to a critical divergence from standard uncertainty sampling.
In classical static models, as $t\to\infty$, the variance $\text{Var}(\theta)\to0$,
which effectively extinguishes the motive to learn. Consequently,
a standard agent eventually becomes a silent scholar again once it
is confident.

In our dynamic framework, assuming a non-trivial environment where
both supporting and contradictory evidence occur with non-zero probability,
the variance converges to a non-zero lower bound, i.e.,
\begin{equation}
\lim_{t\to\infty}\text{Var}(\theta)\ge\delta(\gamma)>0
\end{equation}
This positive floor $\delta(\gamma)$ ensures that the agent's epistemic
hunger is never fully satiated. The agent is mathematically forced
to remain an active participant in the digital commons, not because
it is altruistic, but because its internal belief model constantly
decays toward entropy.

\subsubsection{Maximum Ambiguity\label{subsec:maximum-ambiguity}}

As defined in Section \ref{subsec:persistent-uncertainty}, the agent's
uncertainty is the variance of its belief distribution, $\text{Var}\left(\theta\right)$.
For any given number of total observations $N=\alpha+\beta$, this
variance is maximised when the numerator $\alpha\beta$ is maximised.
This occurs when the agent's pseudo-counts are balanced, i.e., $\alpha=\beta$.
Correspondingly, when $\alpha=\beta$, the agent's expected success
rate is,
\begin{equation}
E[\theta]=\frac{\alpha}{\alpha+\alpha}=0.5.
\end{equation}
In other words, the agent is in a state of maximum uncertainty when
the proposition's effectiveness is ambiguous ($E[\theta]=0.5$). While
traditionally treated as a heuristic, recent theoretical work by Fuchsgruber
et al. \cite{fuchsgruber2025uncertaintyactivelearninggraphs} on graph
active learning proves that acquiring labels for instances with maximal
epistemic uncertainty is mathematically equivalent to maximising the
gain in the posterior probability of the ground truth. Conversely,
their analysis confirms that targeting aleatoric uncertainty yields
sub-optimal performance comparable to random sampling. Thus, by targeting
the peak of $\text{Var}(\theta)$, our agent is formally optimising
its rate of learning.

In other words, the agent is in a state of maximum uncertainty when
the proposition's effectiveness is ambiguous. It is a known property
of this model that this state of maximum prior variance is also the
state from which a single new observation provides the largest expected
reduction in uncertainty (i.e., the highest information gain). When
an agent operating with $E[\theta]\approx0.9$ or $E[\theta]\approx0.1$
is highly certain, a new observation will only minimally shift its
belief. In contrast, an agent with $E[\theta]=0.5$ will experience
a significant belief update regardless of the outcome. 

A full analysis of whether an agent should pursue topics of maximum
uncertainty or a random-selection approach, is beyond the scope of
this paper. This concept, however, is the central focus of uncertainty
sampling in the field of active learning, and we refer interested
readers to the recent literature \cite{Chong_2023,xia2025selectiongenerationsurveyllmbased,doucet2025bridgingdiversityuncertaintyactive}. 

\subsubsection{Adaptability-Certainty Trade-off\label{subsec:Adaptability-Certainty-Trade}}

The forgetting factor $\gamma$ is a critical meta-parameter that
controls the agent's fundamental assumptions about its environment.
Its value dictates the trade-off between the agent's adaptability
(how quickly it responds to new, contradictory evidence) and its certainty
(the baseline level of uncertainty it can ever achieve).

We've established that the effective sample size ($N_{\text{eff}}$)
stabilises at $N_{eq}=\frac{1}{1-\gamma}$ in Section \ref{subsec:persistent-uncertainty}.
This leads to a direct, quantifiable trade-off. When the forgetting
factor $\gamma$ is high, such as $\gamma=0.999$, which corresponds
to $N_{\text{eq}}=1000$, the agent effectively has a long memory.
It places greater trust in its large pool of historical data. As a
result, the agent to become more certain about various propositions,
but slower to adapt. A few new, contradictory pieces of evidence have
little impact and are essentially overwhelmed by the accumulated weight
of the previous 1000 observations.

Conversely, when $\gamma$ is low, such as $\gamma=0.95$, which yields
$N_{\text{eq}}=20$, the agent\textquoteright s effective memory is
short. It trusts primarily the most recent 20 observations. This leads
to a higher baseline level of uncertainty and the agent cannot achieve
high confidence in its beliefs. For the trade-off, the agent has a
strong adaptability. A few new, contradictory observations can quickly
and significantly alter its beliefs.

\subsection{Scalability via Epistemic Caching \label{subsec:Scalability-via-Epistemic}}

While the forgetting factor ($\gamma$) successfully establishes the
motive for continuous interaction, it simultaneously introduces a
computational constraint: an agent cannot maintain active, decaying
belief states for the infinite space of all possible propositions.
However, the very mechanism that creates this constraint also provides
its solution. Since the effective sample size $N_{\text{eff}}$ decays
naturally over time, it functions as an intrinsic epistemic timer,
allowing us to implement a cache eviction policy that is not heuristic,
but probabilistically derived from the agent\textquoteright s own
fading memory.

In real-world deployments, the number of potential propositions is
enormous and often follows a long-tail distribution. Most queries
are unique, however a specific subset of active propositions is frequently
accessed.

Our framework addresses this through dynamic epistemic caching. Similar
to cache management in computer architecture, the agent dynamically
allocates parameters only to propositions in the active head of the
distribution, while relying on the LLM's generic pre-training for
the long tail.

The forgetting factor $\gamma$ naturally governs the eviction policy
for this cache. For any rarely accessed proposition $p_{\text{rare}}$,
the effective sample size $N_{\text{eff}}$ decays over time according
to the recurrence relation derived in Section \ref{subsec:persistent-uncertainty},
i.e., 
\begin{equation}
N_{\text{eff},t}=N_{\text{eff},t-1}\times\gamma+\mathbb{I}_{\text{obs}},
\end{equation}
where $\mathbb{I}_{\text{obs}}$ represents the binary observation
signal, 
\begin{equation}
\mathbb{I}_{\text{obs}}=\begin{cases}
1 & \text{if new evidence is observed at time }t\\
0 & \text{otherwise}.
\end{cases}
\end{equation}
We define a significance threshold $N_{\text{min}}$ to formalise
the eviction condition. A proposition $p_{i}$ is evicted from the
active tracking set and reverted to the LLM's generic pre-training
when $N_{\text{eff},t}<N_{\text{min}}.$

This mechanism functions as a variation of Least Recently Used (LRU)
eviction weighted by epistemic density. Unlike standard LRU, which
relies solely on recency, our approach preferentially retains high-confidence
propositions (high $N_{\text{eff}}$), ensuring that computational
resources remain concentrated on high-utility, frequently accessed
knowledge.

\section{Simulation Results\label{sec:Simulation-Results}}

To validate the theoretical properties of our Beta-Bernoulli model
with forgetting factor, we conducted simulations. The goals are to
(a) demonstrate the necessity of the forgetting factor ($\gamma$)
in a non-stationary environment, validating the adaptability-certainty
trade-off, and (b) evaluate the efficacy of uncertainty sampling in
a standard uniform environment, and (c) stress-test the epistemic
caching mechanism under realistic heterogeneous (Zipfian) access patterns.

\subsection{Experiment Setup}

We simulate a digital commons environment containing $k=100$ independent
propositions. To model the non-stationary nature of the digital commons
(e.g., evolving public consensus), we introduce a consensus shift.
For time steps $t=1...500$, the ground truth is $\theta^{*}=0.8$
(strong consensus). At $t=501$, we simulate a paradigm shift, where
the ground truth permanently flips to $\theta^{*}=0.2$ (contradiction).

We define two environmental access conditions:
\begin{enumerate}
\item Uniform Access: The agent selects queries from a uniform distribution
over the $k$ propositions.
\item Heterogeneous Access: To mirror real-world long-tail distributions,
the agent samples propositions according to a Zipfian (power law)
distribution ($s=1.1$). This creates a scenario where a small subset
of head propositions is frequently accessed, while the majority are
rarely queried.
\end{enumerate}

\subsection{Experiment 1: Validation of the Adaptability-Certainty Trade-off
\label{subsec:experiment1}}

This experiment validates the adaptability-certainty trade-off (Section
\ref{subsec:Adaptability-Certainty-Trade} ). To strictly isolate
the impact of the forgetting factor ($\gamma$) from the query strategy,
we employ a naive random sampling strategy across all three agents
in a uniform environment: a static agent ($\gamma=1.0$), a high-$\gamma$
agent ($\gamma=0.999$), and a low-$\gamma$ agent ($\gamma=0.95$).

\begin{figure}

\begin{centering}
\includegraphics[width=1\columnwidth]{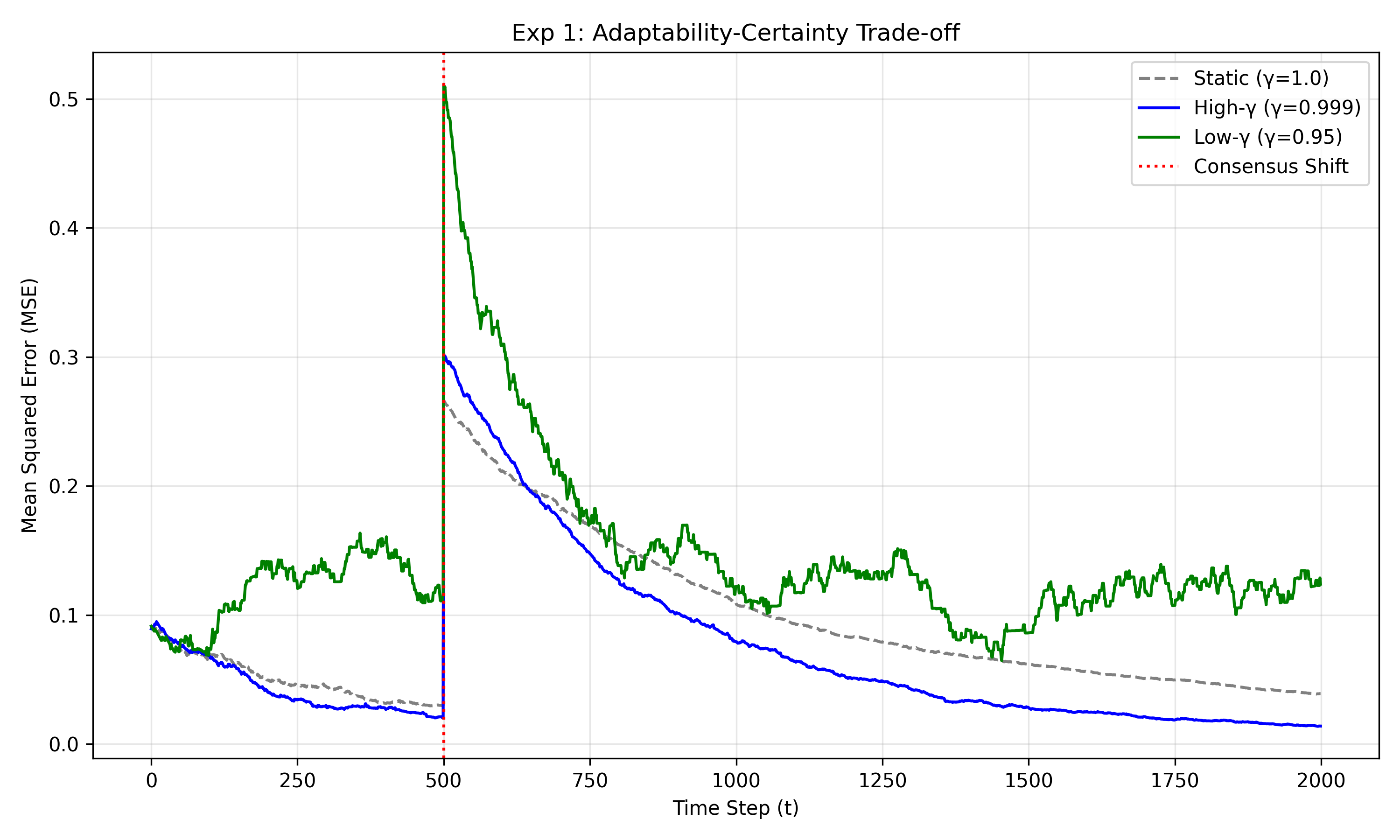}
\par\end{centering}
\caption{Adaptability trade-off using random sampling. The low-$\gamma$ agent
(green) suffers from high noise but adapts rapidly to the shift. The
high-$\gamma$ agent (blue) offers stability but exhibits significant
inertia, adapting slower than the static agent (grey) due to its larger
effective memory horizon ($N_{\text{eq}}=1000$ vs $t=500$). \label{fig:adaptability-centainty-trade-off}}
\end{figure}

As shown in Figure \ref{fig:adaptability-centainty-trade-off}, the
results illustrate the tension between noise and inertia.
\begin{itemize}
\item The Cost of Stability: The high-$\gamma$ agent (blue) achieves the
lowest baseline error before the shift, effectively filtering noise.
However, it exhibits significant inertia after the shift at $t=500$.
Notably, it recovers slower than the static agent (grey) in this specific
window. This is because the high-$\gamma$ agent maintains an effective
memory of $N_{\text{eq}}\approx1000$ steps, whereas the static agent
had only accumulated 500 steps of history at the time of the shift.
This confirms that while high stability is beneficial for precision,
it creates a heavy prior that resists new evidence.
\item The Benefit of Forgetting: The low-$\gamma$ agent (green) maintains
a higher noise floor ($\approx0.1$) but adapts rapidly. Its steep
recovery slope after $t=500$ confirms that a lower forgetting factor
is essential for survival in highly volatile environments, allowing
the agent to quickly discard obsolete beliefs.
\end{itemize}

\subsection{Experiment 2: Efficacy in Uniform Environment \label{subsec:experiment2}}

We compare the random sampling agent against the uncertainty sampling
agent in a uniform access environment with $\gamma=0.999$.

As shown in Figure \ref{fig:efficacy-uniform-env}, the uncertainty
sampling agent (orange) demonstrates superior learning efficiency
during the stable phase ($t<500$), driving MSE significantly lower
than the random agent. However, the results highlight a critical re-calibration
penalty. Immediately following the consensus shift ($t=501$), the
uncertainty agent spikes to a higher error than random sampling and
lags behind. This occurs because the uncertainty agent, having built
high confidence in the old regime, must systematically dismantle its
strong beliefs. It aggressively targets the ambiguity of the transition
($\text{E}\left[\theta\right]\approx0.5$), effectively over-thinking
the shift, while the random agent benefits from broader exploration.

\begin{figure}
\begin{centering}
\includegraphics[width=1\columnwidth]{images/experimentB-2}
\par\end{centering}
\caption{Strategy comparison (Uniform). Uncertainty sampling (orange) minimises
error efficiently in stable regimes but incurs a severe re-calibration
penalty after the shift ($t=500$), temporarily lagging behind random
sampling. \label{fig:efficacy-uniform-env}}
\end{figure}

\subsection{Experiment 3: Robustness in Heterogeneous Environment \label{subsec:experiment3}}

Finally, we test the agents in the heterogeneous (Zipfian) environment
to validate the epistemic caching mechanism (Section \ref{subsec:Scalability-via-Epistemic}).

\begin{figure}
\begin{centering}
\includegraphics[width=1\columnwidth]{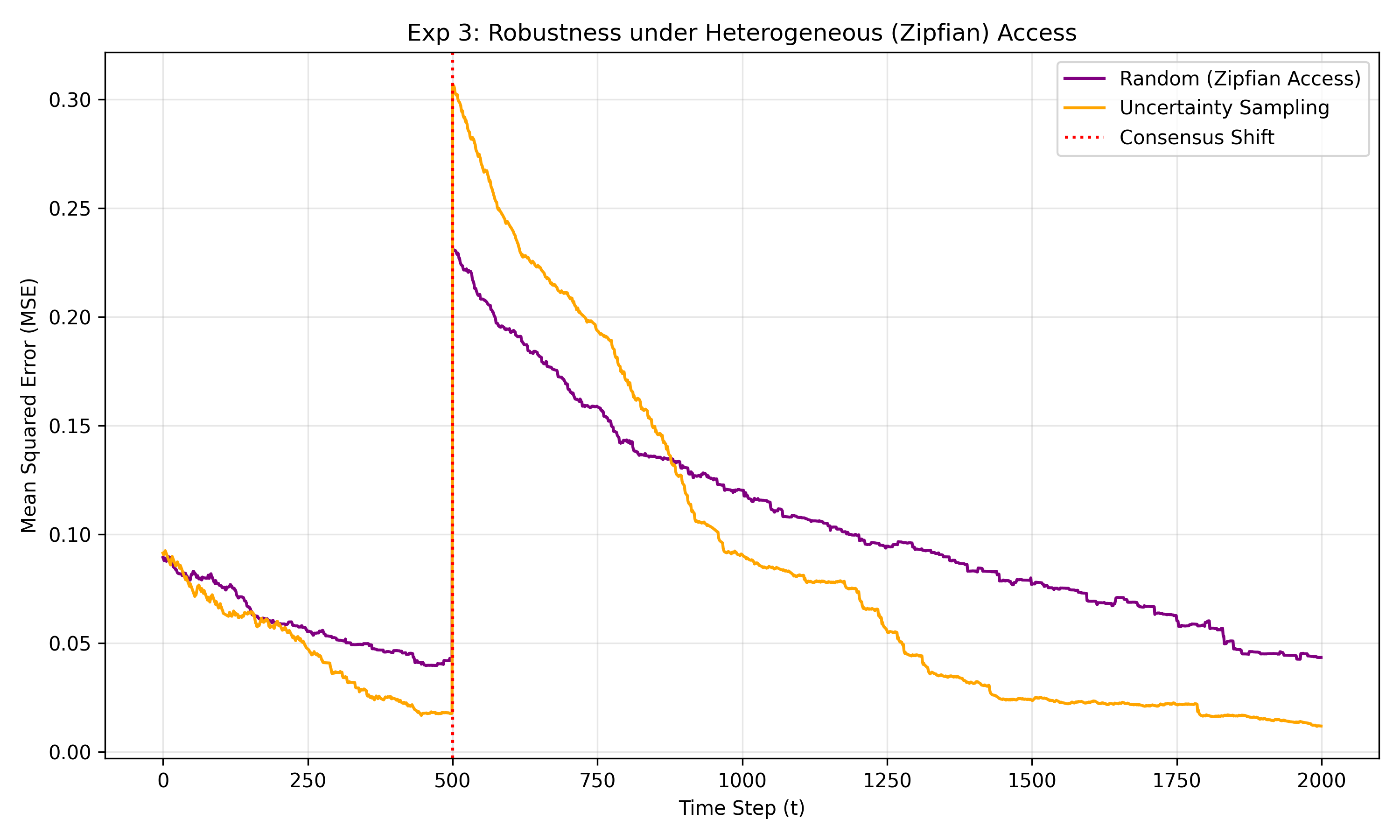}
\par\end{centering}
\caption{Strategy comparison (Zipfian). Random sampling (purple) fails to learn
effectively in a long-tail environment. Uncertainty sampling (orange)
demonstrates robustness; despite the initial re-calibration spike,
it successfully converges to a lower error, proving its ability to
prioritize the active head of the distribution. \label{fig:robustness-heterogeneous-env}}
\end{figure}

As shown in Figure \ref{fig:robustness-heterogeneous-env}, the distinction
between strategies becomes critical.
\begin{itemize}
\item Inefficiency of Randomness: The random agent (purple) performs poorly
throughout, with a slow learning curve. In a long-tail distribution,
random sampling wastes the majority of its query budget on irrelevant,
low-frequency propositions, failing to maintain the active head of
knowledge.
\item Long-Term Dominance of Uncertainty Sampling: The uncertainty sampling
agent (orange) initially suffers the same re-calibration penalty observed
in the experiment in Section \ref{subsec:experiment2}. However, unlike
the random agent, it successfully learns. By targeting variance, the
agent automatically concentrates its resources on the frequent, high-variance
propositions. Around $t=900$, it crosses below the random agent and
continues to drive error down. This confirms that in realistic, skewed-access
environments, uncertainty sampling is the only viable strategy for
maintaining a high-fidelity belief state over time.
\end{itemize}

\section{Discussion: Implications for Alignment and Robustness \label{sec:Discussion}}

Our experiments demonstrate that uncertainty sampling effectively
governs an agent's immediate query strategy. However, the probabilistic
belief states ($\alpha,\beta$) generated by this process possess
utility that extends beyond real-time interaction. In this section,
we discuss how these internal uncertainty metrics can be repurposed
to enhance long-term model alignment and system robustness in non-stationary
environments.

\subsection{Implications for Model Alignment}

Beyond immediate query strategies, the agent's accumulated belief
states offer a verifiable signal for long-term alignment. We identify
the following key mechanisms for this integration.

\subsubsection{Dynamic SFT Filtering}

Propositions with high confidence ($\text{Var}(\theta)\to0$) and
high success rates effectively identify the gold standard subset of
an agent's experiences. These can be extracted to create high-quality,
autonomously curated datasets for supervised fine-tuning (SFT), aligning
with recent findings that data quality outweighs quantity for model
robustness \cite{zhou2023limaalignment}.

\subsubsection{\LyXZeroWidthSpace Epistemic Reward Signals}

The internal belief state can serve as a dense reward signal for reinforcement
learning from human feedback (RLHF). By penalizing reasoning paths
that contradict high-confidence internal beliefs, the agent can be
incentivized to maintain consistency, bridging the gap between inference-time
reasoning and training-time alignment \cite{zhai2023uncertaintypenalizedreinforcementlearninghuman}.
Beyond providing reward signals, these accumulated beliefs can be
synthesized into the agent's core architecture through a process of
continuous distillation.

\subsubsection{Continuous Distillation}

Rather than relying solely on external memory (the portfolio $P$),
the agent can periodically distill its accumulated belief state into
the LLM's weights. By training on the high-certainty active head of
its portfolio (as defined in Section \ref{subsec:Scalability-via-Epistemic}),
the agent consolidates transient, external observations into permanent,
internal knowledge \cite{west2022symbolicknowledgedistillationgeneral},
effectively solving the catastrophic forgetting problem by using the
Beta-Bernoulli parameters as a stable buffer.

\subsection{Mitigating Re-calibration Latency}

Our simulations in Section \ref{subsec:experiment2} identified a
significant re-calibration penalty, where uncertainty sampling lagged
behind random sampling following the consensus shift. To mitigate
this in deployment, future architectures could employ a surprisal
reset mechanism. If the prediction error (KL-divergence) exceeds a
critical threshold, the agent could temporarily reset the effective
sample size ($N_{\text{eff}}$\LyXZeroWidthSpace ) for that proposition.
This would instantly restore maximum plasticity, allowing the agent
to bypass the inertia observed in Figure \ref{fig:efficacy-uniform-env}
and rapidly adapt to the new ground truth.

\subsection{Model Limitations and Real-World Knowledge Structures}

While the proposed framework establishes a formal foundation for the
non-altruistic motive of epistemic agents, it relies on two simplifying
assumptions that warrant further discussion: proposition independence
and binary feedback signals. This section discusses extending the
model to real-world complexities, including graph-structured dependencies,
multi-categorical observations, and hierarchical epistemic caching.

\subsubsection{Inter-Proposition Dependencies}

In this study, we treat the knowledge base $P$ as a portfolio of
$k$ independent propositions. However, real-world knowledge is often
structured hierarchically or as a graph (e.g., knowledge graphs),
where the truth value of one proposition is highly correlated with
others. In such systems, an observation $y_{t}$ for proposition $p_{i}$
provides a leakage of information that should theoretically reduce
the epistemic uncertainty $(\text{Var}(\theta))$ of related propositions
$p_{j}$. Future work could extend our Beta-Bernoulli update rule
to a graph-based Bayesian framework. In this configuration, updates
to $\alpha$ and $\beta$ would propagate through the graph edges
based on semantic similarity or logical entailment, potentially accelerating
the agent's convergence in complex, multi-layered domains.

\subsubsection{Beyond Binary Observations}

Our model currently defines the observation $y_{t}\in\{0,1\}$ as
a discrete indicator of success or failure. While this unit of evidence
is mathematically convenient for deriving the equilibrium sample size
$N_{\text{eq}}$, contemporary digital feedback is often nuanced and
non-binary. For instance, feedback from the digital commons may take
the form of confidence scores or partial critiques. Our framework
is inherently compatible with such soft labels. Instead of adding
an integer value to the pseudo-counts, the update rule in Eq. \ref{eq:alpha-update}
and \ref{eq:beta-update} can incorporate a fractional value $s\in[0,1]$,
where $s$ represents the degree of supporting evidence.

\subsubsection{Epistemic Density in Hierarchies}

The introduction of dependencies would also enhance our proposed epistemic
caching. Rather than evicting individual propositions based solely
on their specific $N_{\text{eff}}$, a hierarchical model could prioritise
the retention of foundational propositions that support many leaf
nodes. This would ensure that an agent\textquoteright s limited computational
cache is reserved for the most structurally significant knowledge,
further optimising the trade-off between adaptability and certainty
in high-velocity environments.

\section{Conclusion\label{sec:Conclusion}}

Contemporary AI agents function as silent scholars, passively consuming
knowledge without active engagement. We argue their evolution into
epistemic agents will be driven by a formal, non-altruistic motive
to learn, compelling them to interact with the digital commons to
verify truth propositions.

To ground this, we proposed a probabilistic framework using a Beta-Bernoulli
model with a forgetting factor ($\gamma$). Our analysis demonstrates
that this forgetting factor acts as the critical link between aleatoric
volatility and epistemic maintenance. By forcing belief distributions
to decay, the framework prevents epistemic uncertainty from vanishing,
thereby establishing a permanent, homeostatic motive for continuous
engagement.

Furthermore, our simulations empirically validate the mechanism of
epistemic caching. We demonstrated that under realistic, heterogeneous
(Zipfian) access patterns, the forgetting factor naturally reallocates
computational resources to the non-stationary active head of the distribution,
allowing the agent to outperform random baselines significantly in
long-tail environments. While this study serves as a foundational
proof-of-concept assuming proposition independence and binary feedback,
it paves the way for more sophisticated architectures. Finally, we
highlighted the framework's potential to close the loop between inference
and training, proposing that the agent's accumulated belief state
can serve as a continuously curated dataset for autonomous model alignment
and the mitigation of catastrophic forgetting.

In conclusion, we reframe public contribution not as altruism, but
as optimal active learning, a formally-justified, rational strategy
for an agent to accelerate its own learning.

\bibliographystyle{plain}
\bibliography{bibliography}

\end{document}